\documentclass{article}
\usepackage{kr}

\usepackage{times}

\usepackage{soul}
\usepackage{url}
\usepackage[hidelinks]{hyperref}
\usepackage[utf8]{inputenc}
\usepackage[small]{caption}
\usepackage{graphicx}
\usepackage{amsmath}
\usepackage{amssymb}
\usepackage{booktabs}
\usepackage{tikz}
\usepackage{xspace}
\usepackage[ruled]{algorithm2e}
\usepackage{caption}
\usepackage{subcaption}
\usepackage{cancel}
\usepackage[autolanguage,np]{numprint}
\newcommand{\topconf}[1]{\mathrm{TopSol}_{#1}}
\newcommand{\topval}[1]{\mathrm{TopVal}_{#1}}
\newcommand{\transfv}[1]{\mathrm{TopTra}_{#1}}
\newcommand{\valida}{\mathrm{ValidA}}
\newcommand{\validv}{\mathrm{ValidV}}

\newcommand{\mm}[1]{{\small{\textsf{#1}}}}

\newcommand{\vt}{\mathrm{1}}
\newcommand{\vf}{\mathrm{0}}

\newcommand{\lit}{\mathrm{Lit}}

\def\Dnnf{{\tt DNNF}}
\def\dDnnf{{\tt d-DNNF}}
\def\cnf{{\tt CNF}}

\def\obdd{{\tt OBDD}}
\def\fbdd{{\tt FBDD}}

\newcommand{\eat}[1]{}

\newcommand{\dfour}{\mm{d4}}

\newcommand{\adnan}{\mm{c2d}}
\newcommand{\Dsharp}{\mm{Dsharp}}

\newtheorem{definition}{Definition}
\newtheorem{theorem}{Theorem}

\newtheorem{example}{Example}

\title{Pseudo Polynomial-Time Top-$k$ Algorithms for \dDnnf\ Circuits}

\author{
Pierre Bourhis$^1$\and
Laurence Duchien$^1$\and
Jérémie Dusart$^1$ \and
Emmanuel Lonca$^2$\and\\
Pierre Marquis$^{2,3}$ \and
Clément Quinton$^1$\\
\affiliations
$^1$University of Lille, CNRS, INRIA, CRIStAL\\
$^2$Univ. Artois, CNRS, CRIL\\
$^3$Institut Universitaire de France\\
\emails
\{pierre.bourhis, laurence.duchien, jeremie.dusart, clement.quinton\}@univ-lille.fr,\\
\{lonca, marquis\}@cril.fr
}

\hypersetup{draft}
\begin{document}

\maketitle

\begin{abstract}
We are interested in computing $k$ most preferred models of a given \dDnnf\ circuit $C$,
where the preference relation is based on an algebraic structure called 
a monotone, totally ordered, semigroup $(K, \otimes, <)$. In our setting, every literal in $C$ has a value 
in $K$ and the value of an assignment is an element of $K$ obtained by aggregating using $\otimes$ the values of the
corresponding literals. We present an algorithm that computes $k$ models of $C$ among those having the largest values 
w.r.t. $<$, and show that this algorithm runs in time polynomial in $k$ and in the size of $C$. We also 
present a pseudo polynomial-time algorithm for 
deriving the top-$k$ values that can be reached, provided that an additional (but not very demanding) requirement on the semigroup
is satisfied.
Under the same assumption, we present a pseudo polynomial-time algorithm that 
transforms $C$ into a \dDnnf\ circuit $C’$ satisfied exactly by the models of $C$ having a value among
the top-$k$ ones. Finally, focusing on the semigroup $(\mathbb{N}, +, <)$, we compare on a large number of instances 
the performances of our compilation-based algorithm for computing $k$ top solutions with 
those of an algorithm tackling the same problem, but based on a partial weighted MaxSAT solver.
\end{abstract}

\section{Introduction}

In this paper, we are interested in \emph{optimization problems under compiled constraints}.
Roughly, the goal is to derive most preferred solutions among the feasible ones, where the set of
feasible solutions is of combinatorial nature and represented implicitly as valid assignments,
i.e., those truth assignments satisfying some given constraints. Such optimization questions are key issues in a number of applications 
about configuration, recommendation, and e-commerce (see e.g., \cite{8572588,DBLP:journals/aim/JannachPRZ21,DBLP:reference/sp/2015rsh,DBLP:conf/edbt/KhabbazL11}).

Unlike the preference relation at hand that is user-specific, the set of constraints representing the 
valid assignments is typically independent of the user, so that it does not often change.
In such a case, taking advantage of a knowledge compilation
  approach can be useful, since compiling the constraints during an
  offline phase may allow polynomial-time algorithms for optimization
  tasks, whilst getting a single optimal solution already is {\sf NP}-hard 
  when no assumptions are made on the representations of the constraints. The discrepancy
  between the two approaches is amplified when computing multiple best
  solutions. Indeed, in practice, the computation of several best solutions for a set of
  uncompiled constraints generally requires successive calls to a {\sf NP}
  oracle (see e.g., \cite{DBLP:conf/pkdd/JabbourSS13}), while one can expect
  polynomial-time algorithms for this task too when the constraints have been compiled first.

When dealing with propositional constraints, the language of deterministic, decomposable Negation Normal Form circuits (\dDnnf)
\cite{Darwiche01} appears as a valuable language for compiling constraints because it supports in polynomial time a
number of queries and transformations that are {\sf NP}-hard in general \cite{Darwiche_2002}. Among them are queries
and transformations about optimization. Thus, \cite{DarwicheMarquis04} has shown how to derive in polynomial time
a most preferred, yet feasible solution 
where the set of feasible solutions is the set of models of a given 
\dDnnf\ circuit and the value of a solution is the sum of the weights (numbers) associated with the literals it satisfies. The authors have also 
presented a polynomial-time transformation that returns a \dDnnf\ circuit whose models are precisely the preferred, feasible
solutions of the \dDnnf\ circuit one started with. 

Such an approach has been extended to a much more general, algebraic model counting
setting in \cite{DBLP:journals/japll/KimmigBR17}. The extension that has been achieved is threefold: first, the value of a solution
is not necessarily a number, but an element of an abstract set, the carrier $K$ of an algebraic structure called a commutative semiring;
then, the aggregation operator used to define the value of a solution is not restricted to summation, but can be any abstract binary operator $\otimes$
over $K$; finally, the authors take advantage of an additional aggregation operator, $\oplus$, which is not necessarily equal to $\mathit{max}$ or
$\mathit{min}$. The goal is to compute the algebraic model count of a given \dDnnf\ circuit $C$, defined as the aggregation using $\oplus$
of the values of all models of $C$, where the value of a solution is the aggregation using $\otimes$ of the values of the literals satisfied by the
solution (such values are elements of $K$). Algebraic model counting generalizes a number of problems of interest, including satisfiability (SAT), (possibly weighted) model counting (\#SAT-WMC), 
and probabilistic inference (PROB) (see Theorem 1 in \cite{DBLP:journals/japll/KimmigBR17}).

Following \cite{DBLP:journals/japll/KimmigBR17}, we extend the approach to optimization under 
\dDnnf\ constraints considered in \cite{DarwicheMarquis04} but the generalization made here relies on a different perspective, as reflected by the queries and transformation we focus on. Whilst \cite{DarwicheMarquis04} aims to compute a single,
most preferred solution (and the corresponding value), we are interested in computing $k$ most preferred models 
of a given \dDnnf\ circuit $C$, where $k$ is a preset bound given by the user. The returned assignments must be valid and
their values must be among the largest possible ones, i.e., for any top-$k$ assignment $\omega$, there cannot exist $k$ (or more) valid assignments
having a value strictly greater than the one of $\omega$.

Considering top-$k$ solutions is important to handle situations when the user is not satisfied by the top-$1$ solution that is provided (maybe he/she would 
finally prefer another solution reaching the top value, or even a solution with value slightly smaller than the value of a top-$1$ solution).  
We are also interested in computing the $k$ most preferred values, thus extending the issue of computing a top-$1$ solution and 
the top-$1$ value as considered in \cite{DarwicheMarquis04} 
to the computation of top-$k$ solutions and top-$k$ values. Finally, we investigate the corresponding transformation problem. 

As \cite{DBLP:journals/japll/KimmigBR17}, we  consider a more general algebraic setting than the one in \cite{DarwicheMarquis04} 
where, implicitly, $\otimes$ is the summation operator and $K$ is the set of real numbers. We focus here on an algebraic structure called 
a \emph{monotone, totally ordered, semigroup} $(K, \otimes, <)$. Notwithstanding the $\oplus$ operator 
(which is implicitly $\mathit{max}$ in our case) the structure used is less demanding than commutative semirings;
especially, in the general case, the existence of a neutral element for $\oplus = \mathit{max}$ (i.e., a least element in $K$ w.r.t. $<$)
that is an annihilator for $\otimes$ is not required. Given a mapping $\nu$ associating with every literal $\ell$ of $C$
an element of $K$, the value $\nu(\omega)$ of an assignment $\omega$ is defined as 
$\nu(\omega) = \otimes_{\ell \in \mathit{Var}(C)  \mid \omega \models \ell} \nu(\ell)$.
On this ground, a top-$k$ value of $C$ given $(K, \otimes, <)$ and $\nu$ 
is one of the $k$-largest values $v$ of $K$ w.r.t. $<$ such that $v = \nu(\omega)$ is the value of a valid assignment $\omega$ of $C$.
A top-$k$ solution of $C$ given $(K, \otimes, <)$ and $\nu$ is a model $\omega$ of $C$ such that there is 
strictly less than $k$ valid assignments
of $C$ having a value strictly greater w.r.t. $<$ than $\nu(\omega)$.

Our contribution is as follows. We first present an algorithm that computes top-$k$ solutions of $C$ given $(K, \otimes, <)$ and $\nu$
and that runs in time polynomial in $k$ and in the size of the \dDnnf\ circuit $C$. We also present a pseudo polynomial-time algorithm for 
deriving the top-$k$ values of $C$ given $(K, \otimes, <)$ and $\nu$, provided that an additional (but not very demanding) 
requirement on the semigroup $(K, \otimes, <)$, namely almost strict monotony, 
is satisfied. 
Under the same assumption, 
we present a pseudo polynomial-time algorithm that 
transforms $C$ into a \dDnnf\ circuit $C’$ satisfied exactly by the models of $C$ having a value among
the top-$k$ values of $C$ given $(K, \otimes, <)$ and $\nu$. 
Whenever $k$ is small enough so that it can be considered as bounded by a constant (which is a reasonable assumption in practice), 
each of our top-$k$ algorithms runs in time linear in the size of the \dDnnf\ circuit $C$. 
Finally, focusing on the semigroup $(\mathbb{N}, +, <)$, we present the results of an empirical comparison  
of our compilation-based algorithm for computing top-$k$ solutions with an algorithm tackling the same problem,
but based on the partial weighted MaxSAT solver {\tt MaxHS} \cite{DBLP:conf/cp/DaviesB11,davies2013solving,10.1007/978-3-642-39071-5_13,10.1007/978-3-642-40627-0_21}. The obtained results show that 
in practice taking advantage of the compilation-based algorithm makes sense for many instances.

The rest of the paper is organized as follows. We give some preliminaries in Section \ref{sec:prelim}. Then we
present our algorithm for computing $k$ top solutions of a \dDnnf\ circuit in Section \ref{sec:topsol} and our algorithm for computing 
its top-$k$ values in Section \ref{sec:topval}. Our algorithm for achieving the top-$k$ transformation of a  \dDnnf\ circuit is
presented in Section \ref{sec:toptra}. The results of the empirical evaluation are given in Section \ref{sec:expe}. 
Possible extensions of our results are briefly discussed in Section \ref{sec:disc}. Finally,
Section \ref{sec:concl} concludes the paper.
The code of our algorithms can be found at \url{https://gitlab.inria.fr/jdusart/winston} while the data used
in the experiments are available from \url{https://gitlab.inria.fr/jdusart/knowledge-compilation-xp}.

\section{Preliminaries}\label{sec:prelim}


Let $X$ be a set of propositional variables.  
The set of literals over $X$ is the union of $X$ with the set of negated variables over $X$, and it and denoted by $\lit(X)$.
An \emph{assignment} $\omega$ is a mapping from $X$ to $\{\vt,\vf\}$.
A \emph{Boolean  function} $f$ over $X$ is a mapping from the assignments over $X$ to  $\{\vt,\vf\}$. 
An assignment $\omega$ such that $f(\omega) = \vt$ is called a \emph{valid} assignment, or a model of $f$.
The set of valid assignments for $f$ is noted $\valida(f)$.
%

\paragraph{\dDnnf\ circuits.}
Circuits 
are convenient representations of Boolean functions. 
A deterministic, Decomposable Negation Normal Form (\dDnnf) circuit \cite{Darwiche01} is a 
directed acyclic graph (DAG) 
where internal nodes are labelled by connectives in $\{\wedge, \vee\}$  and leaves are labelled by literals from $\lit(X)$ or Boolean constants.
The two main properties of \dDnnf\ circuits are that the sets of variables appearing in the subcircuits of any $\wedge$ node 
are pairwise disjoint (decomposability) and the valid assignments of the subcircuits of any $\vee$ node are pairwise disjoint (determinism).
Figure~\ref{fig:ddnnf:ddnnf} gives an example of a \dDnnf\ circuit.

\newcommand{\figTupleAnd}{,}
\newcommand{\figScale}{.85}

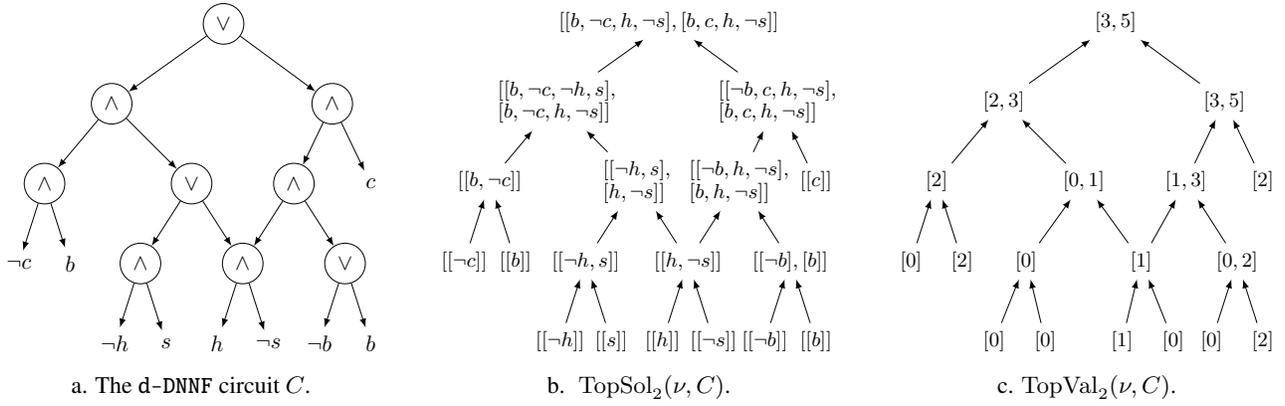
\begin{figure*}[t]
  \centering
  \begin{subfigure}[b]{0.33\textwidth}
    \centering
    \scalebox{\figScale}{
    \begin{tikzpicture}
      \tikzstyle{internalAndNode}=[draw, circle]
      \tikzstyle{internalOrNode}=[draw, circle]
      \tikzstyle{leafNode}=[]
      \tikzstyle{link}=[->,>=latex]
      
      \node[internalOrNode] (or1) at (.5,0) {$\vee$};
      \node[internalAndNode] (and1) at (-1.25,-1.25) {$\wedge$};
      \draw[link] (or1) -- (and1);
      \node[internalAndNode] (and5) at (-2.3,-2.5) {$\wedge$};
      \draw[link] (and1) -- (and5);
      \node[leafNode] (nc) at (-2.7,-3.75) {$\neg c$};
      \draw[link] (and5) -- (nc);
      \node[leafNode] (b) at (-1.9,-3.75) {$b$};
      \draw[link] (and5) -- (b);
      \node[internalOrNode] (or2) at (0,-2.5) {$\vee$};
      \draw[link] (and1) -- (or2);
      \node[internalAndNode] (and2) at (-.8,-3.75) {$\wedge$};
      \draw[link] (or2) -- (and2);
      \node[leafNode] (nh) at (-1.2,-5) {$\neg h$};
      \draw[link] (and2) -- (nh);
      \node[leafNode] (s) at (-.4,-5) {$s$};
      \draw[link] (and2) -- (s);
      \node[internalAndNode] (and3) at (.8,-3.75) {$\wedge$};
      \draw[link] (or2) -- (and3);
      \node[leafNode] (h) at (.4,-5) {$h$};
      \draw[link] (and3) -- (h);
      \node[leafNode] (ns) at (1.2,-5) {$\neg s$};
      \draw[link] (and3) -- (ns);
      \node[internalAndNode] (and4) at (1.6,-2.5) {$\wedge$};
      \node[internalAndNode] (and6) at (2.2,-1.25) {$\wedge$};
      \draw[link] (or1) -- (and6);
      \draw[link] (and4) -- (and3);
      \node[leafNode] (c) at (2.8, -2.5) {$c$};
      \draw[link] (and6) -- (c);
      \draw[link] (and6) -- (and4);
      \node[internalOrNode] (or3) at (2.4,-3.75) {$\vee$};
      \draw[link] (and4) -- (or3);
      \node[leafNode] (nb2) at (2,-5) {$\neg b$};
      \node[leafNode] (b2) at (2.8,-5) {$b$};
      \draw[link] (or3) -- (nb2);
      \draw[link] (or3) -- (b2);
    \end{tikzpicture}
    }
    \caption{\label{fig:ddnnf:ddnnf}The \dDnnf\ circuit $C$.}
  \end{subfigure}
  \begin{subfigure}[b]{0.33\textwidth}
    \centering
    \scalebox{\figScale}{
    \begin{tikzpicture}
      \tikzstyle{internalAndNode}=[align=left, font=\small]
      \tikzstyle{internalOrNode}=[align=left, font=\small]
      \tikzstyle{leafNode}=[align=left, font=\small]
      \tikzstyle{link}=[<-,>=latex]
      
      \node[internalOrNode] (or1) at (.5,0) {$[[b \figTupleAnd \neg c \figTupleAnd h \figTupleAnd \neg s], [b \figTupleAnd c \figTupleAnd h \figTupleAnd \neg s]]$};
      \node[internalAndNode] (and1) at (-1.25,-1.25) {$[[b \figTupleAnd \neg c \figTupleAnd \neg h \figTupleAnd s],$\\$[b \figTupleAnd \neg c \figTupleAnd h \figTupleAnd \neg s]]$};
      \draw[link] (or1) -- (and1);
      \node[internalAndNode] (and5) at (-2.3,-2.5) {$[[b \figTupleAnd \neg c]]$};
      \draw[link] (and1) -- (and5);
      \node[leafNode] (nc) at (-2.7, -3.75) {$[[\neg c]]$};
      \draw[link] (and5) -- (nc);
      \node[leafNode] (b) at (-1.9,-3.75) {$[[b]]$};
      \draw[link] (and5) -- (b);
      \node[internalOrNode] (or2) at (0,-2.5) {$[[\neg h \figTupleAnd s],$\\$[h \figTupleAnd \neg s]]$};
      \draw[link] (and1) -- (or2);
      \node[internalAndNode] (and2) at (-.8,-3.75) {$[[\neg h \figTupleAnd s]]$};
      \draw[link] (or2) -- (and2);
      \node[leafNode] (nh) at (-1.2,-5) {$[[\neg h]]$};
      \draw[link] (and2) -- (nh);
      \node[leafNode] (s) at (-.4,-5) {$[[s]]$};
      \draw[link] (and2) -- (s);
      \node[internalAndNode] (and3) at (.8,-3.75) {$[[h \figTupleAnd \neg s]]$};
      \draw[link] (or2) -- (and3);
      \node[leafNode] (h) at (.4,-5) {$[[h]]$};
      \draw[link] (and3) -- (h);
      \node[leafNode] (ns) at (1.2,-5) {$[[\neg s]]$};
      \draw[link] (and3) -- (ns);
      \node[internalAndNode] (and4) at (1.6,-2.5) {$[[\neg b \figTupleAnd h \figTupleAnd \neg s],$\\$[b \figTupleAnd h \figTupleAnd \neg s]]$};
      \node[internalAndNode] (and6) at (2.2,-1.25) {$[[\neg b \figTupleAnd c \figTupleAnd h \figTupleAnd \neg s],$\\$[b \figTupleAnd c \figTupleAnd h \figTupleAnd \neg s]]$};
      \draw[link] (or1) -- (and6);
      \draw[link] (and4) -- (and3);
      \node[leafNode] (c) at (2.8, -2.5) {$[[c]]$};
      \draw[link] (and6) -- (c);
      \draw[link] (and6) -- (and4);
      \node[internalOrNode] (or3) at (2.4,-3.75) {$[[\neg b], [b]]$};
      \draw[link] (and4) -- (or3);
      \node[leafNode] (nb2) at (2,-5) {$[[\neg b]]$};
      \node[leafNode] (b2) at (2.8,-5) {$[[b]]$};
      \draw[link] (or3) -- (nb2);
      \draw[link] (or3) -- (b2);
    \end{tikzpicture}
    }
    \caption{\label{fig:ddnnf:topkconf} $\topconf{2}(\nu,C)$.}
  \end{subfigure}
  \begin{subfigure}[b]{0.33\textwidth}
    \centering
    \scalebox{\figScale}{
    \begin{tikzpicture}
      \tikzstyle{internalAndNode}=[align=left, font=\small]
      \tikzstyle{internalOrNode}=[align=left, font=\small]
      \tikzstyle{leafNode}=[align=left, font=\small]
      \tikzstyle{link}=[<-,>=latex]
      
      \node[internalOrNode] (or1) at (.5,0) {$[3,5]$};
      \node[internalAndNode] (and1) at (-1.25,-1.25) {$[2,3]$};
      \draw[link] (or1) -- (and1);
      \node[internalAndNode] (and5) at (-2.3,-2.5) {$[2]$};
      \draw[link] (and1) -- (and5);
      \node[leafNode] (nc) at (-2.7, -3.75) {$[0]$};
      \draw[link] (and5) -- (nc);
      \node[leafNode] (b) at (-1.9, -3.75) {$[2]$};
      \draw[link] (and5) -- (b);
      \node[internalOrNode] (or2) at (0,-2.5) {$[0,1]$};
      \draw[link] (and1) -- (or2);
      \node[internalAndNode] (and2) at (-.9,-3.75) {$[0]$};
      \draw[link] (or2) -- (and2);
      \node[leafNode] (nh) at (-1.4,-5) {$[0]$};
      \draw[link] (and2) -- (nh);
      \node[leafNode] (s) at (-.6,-5) {$[0]$};
      \draw[link] (and2) -- (s);
      \node[internalAndNode] (and3) at (.9,-3.75) {$[1]$};
      \draw[link] (or2) -- (and3);
      \node[leafNode] (h) at (.6,-5) {$[1]$};
      \draw[link] (and3) -- (h);
      \node[leafNode] (ns) at (1.4,-5) {$[0]$};
      \draw[link] (and3) -- (ns);
      \node[internalAndNode] (and4) at (1.6,-2.5) {$[1,3]$};
      \node[internalAndNode] (and6) at (2.2,-1.25) {$[3,5]$};
      \draw[link] (or1) -- (and6);
      \draw[link] (and4) -- (and3);
      \node[leafNode] (c) at (2.8, -2.5) {$[2]$};
      \draw[link] (and6) -- (c);
      \draw[link] (and6) -- (and4);
      \node[internalOrNode] (or3) at (2.4,-3.75) {$[0,2]$};
      \draw[link] (and4) -- (or3);
      \node[leafNode] (nb2) at (2,-5) {$[0]$};
      \node[leafNode] (b2) at (2.8,-5) {$[2]$};
      \draw[link] (or3) -- (nb2);
      \draw[link] (or3) -- (b2);
    \end{tikzpicture}
    }
    \caption{\label{fig:ddnnf:topkval}$\topval{2}(\nu,C)$.}
  \end{subfigure}
  \caption{\label{fig:ddnnf} Top-$2$ algorithms at work on Example \ref{ex:ex1}.}
\end{figure*}

The language of \dDnnf\ circuits includes several interesting languages as subsets, namely the set of Decision-\Dnnf\ circuits  \cite{OztokDarwiche14} 
(where every $\vee$-node is a decision node), the set of \fbdd\ of free binary decision diagrams \cite{GergovMeinel94}, and its subset \obdd, the set of
ordered binary decision diagrams \cite{Bryant86}.
Interestingly, Decision-\Dnnf\ is more succinct than \fbdd\ and \obdd\ \cite{Razgon16}. 
%
%
Furthermore, the language of \dDnnf\ circuits (and its subsets listed above) supports in polynomial time many queries that are intractable ({\sf NP}-hard) when no restriction is put on the circuit (see \cite{Darwiche_2002,DarwicheMarquis04} for details).

\paragraph{Top-$k$ problems.}\label{sec:topk}
In order to present in formal terms the three main top-$k$ computation problems over \dDnnf\ circuits $C$ we are interested in, we first need to make precise the algebraic structure over which the values of the satisfying assignments of $C$ are evaluated:

\begin{definition}
A \emph{monotone, totally ordered semigroup} is a triple 
$(K,\otimes, <)$ where $K$ is a set that is totally ordered by $<$ (a strict, total ordering), 
$\otimes$ is a binary operator over $K$ that is commutative, associative, and \emph{monotone}, i.e., for any $p,q,r,s \in K$, if $p \leq q$ and $r \leq s$ then $p \otimes r \leq q \otimes s$ (where $x \leq y$ iff $x < y$ or $x = y$). The semigroup is \emph{strictly monotone} iff it is monotone and for any $p,q,r,s \in K$, if $p \leq q$ and $r < s$ then $p \otimes r < q \otimes s$.
$(K, \otimes, <)$ is said to have a \emph{least absorptive element} $a$ whenever $a$ is the least element of $K$ w.r.t. $<$ and $a$ is absorptive for $\otimes$, i.e., 
$\forall x \in K, x \otimes a = a \otimes x = a$.
$(K, \otimes, <)$ is said to be \emph{almost strictly monotone} if either $(K, \otimes, <)$ is strictly monotone or $K$ has a least absorptive element $a$ 
and $(K \setminus\{a\}, \otimes,<)$ is strictly monotone.
\end{definition}

 Clearly enough, whenever $(K, \otimes, <)$ has a least absorptive element $a$, $a$ is neutral for $\oplus = \mathit{max}$. Furthermore,
when $\otimes$ is monotone, it distributes over $\oplus = \mathit{max}$ (which is obviously commutative). Thus, in this case, provided that $\otimes$ has a neutral
element $n$, $(K, \mathit{max}, \otimes, a, n)$ is a commutative semiring. However, the existence of such a neutral element $n$ is not mandatory in our setting.

Here are two examples. $(\mathbb{R}, +, <)$ and $([0,1], \times, <)$ are monotone, totally ordered semigroups. 
In $(\mathbb{R}, +, <)$, the elements of $\mathbb{R}$ may denote utilities and in  $([0,1], \times, <)$, the elements of $[0,1]$ may denote
probabilities. It is easy to check that $(\mathbb{R}, +, <)$ is strictly monotone (just like its restriction $(\mathbb{N}, +, <)$), which implies that it does not have a  least absorptive element, and that $([0,1], \times, <)$ has a least absorptive element (namely, $0$) and it is almost strictly monotone.

Provided a monotone, totally ordered semigroup $(K,$ $\otimes,$ $<)$, evaluating an assignment $\omega$ over $X$ requires to indicate how the literals from $\lit(X)$ are interpreted in $K$. This calls for a notion of value function over $X$ onto $K$:

\begin{definition}
Given a $X$ be a set of propositional variables and a monotone, 
totally ordered semigroup $(K,\otimes, <)$, 
a \emph{value function} $\nu$ over $X$ onto $K$ is a mapping from the literals over $X$ to $K$, 
assigning to each literal $\ell$ an element from $K$ noted $\nu(\ell)$ and called the value of $\ell$. 
\end{definition}

When $\otimes$ is a binary operator over $K$, the value of a literal $\ell$ as given by a value function $\nu$ over $X$ onto $K$ 
can then be extended to the \emph{value of an assignment} $\omega$ over $X$, defined as
the $\otimes$-aggregation of the values of the literals (as given by $\nu$) satisfied by $\omega$ (the order with which they are taken does not
matter as soon as $\otimes$ is commutative and associative). This value is noted $\nu(\omega)$.
We denote by $\validv(C)$ the subset of values from $K$ that are reached by the valid assignments of $C$.
Formally, $\validv(C) = \{\nu(\omega) \mid \omega \in \valida(C)\}$. 

We can now define the three top-$k$ problems of $C$ given $(K, \otimes, <)$ and $\nu$ we focus on:

\begin{definition}
Let $(K,\otimes, <)$ be a monotone, totally ordered semigroup. Let $X$ be a set of variables. Let $\nu$ be a value function over $X$ onto $K$. 
Let $C$ be a Boolean circuit over $X$. 
The problem $\topval{k}(\nu,C)$ consists in computing the set of the $k$ largest values w.r.t. $<$ in $\validv(C)$. 
When $\validv(C)$ contains less than $k$ elements, the set is defined as $\validv(C)$.
\end{definition}

Note that the set of top-$k$ values of $C$ given $(K, \otimes, <)$ and $\nu$ is unique since $K$ is totally ordered by $<$.

To define the problem of generating top-$k$ solutions of $C$ given $(K, \otimes, <)$, one first lifts the notion of value of a 
model $\omega$ of $C$ to the notion of value of a set $S$ of models of $C$, defined as follows:
the value of a set $S = \{\omega_1, \ldots, \omega_m\}$ of models of $C$ is the list of values $(\nu(\omega_{\pi(1)}), \ldots, \nu(\omega_{\pi(m)}))$ 
where $\pi$ is a permutation over $\{1, \ldots, m\}$ such that $\nu(\omega_{\pi(1)}) \geq \ldots \geq \nu(\omega_{\pi(m)}$.
The values of such sets $S$ can then be compared w.r.t. the lexicographic ordering $\succ$ induced by $>$.

\begin{definition}
Let $(K,\otimes, <)$ be a monotone, totally ordered semigroup. Let $X$ be a set of variables. Let $\nu$ be a value function over $X$ onto $K$. 
Let $C$ be a Boolean circuit over $X$. A set $S$ of $k$ models of $C$ is a set of top-$k$ models of $C$ if and only if its value is the maximal value w.r.t. $\succ$ 
reached by sets of $k$ models of $C$. As an exception, when $\valida(C)$ has less than $k$ elements, the (unique) set of top-$k$ models of $C$ is defined as $\valida(C)$. 
Finally, the problem $\topconf{k}(\nu,C)$ consists in computing a set of top-$k$ models of $C$. 
\end{definition}

Unlike the set of top-$k$ values, the set of top-$k$ models of $C$ is not unique in general (for instance, it may exist strictly more than $k$ models $\omega$ of $C$
having a maximal value $\nu(\omega)$).

\begin{definition}
Let $(K,\otimes, <)$ be a monotone, totally ordered semigroup. Let $X$ be a set of variables. Let $\nu$ be a value function over $X$ onto $K$. 
Let $C$ be a Boolean circuit over $X$ from a circuit language $\mathcal{L}$.
The problem $\transfv{k}(\nu,C)$ consists in computing from $C$ a circuit $C'$ in the same language $\mathcal{L}$ as $C$ and whose models are precisely those 
of $\valida(C)$ having a value in $\topval{k}(\nu,C)$.
\end{definition}

Obviously enough, such a circuit $C'$ is not unique in general.  Let us now illustrate on a simple example the top-$k$ problems we consider.

\begin{example}\label{ex:ex1}
Consider the 
\dDnnf\ circuit $C$ over $X = \{b,$ $c,$ $h,$ $s\}$ given at Figure \ref{fig:ddnnf:ddnnf}.   
$C$ encodes an E-Shop security system where at least one payment between {\tt bank transfer (b)} and {\tt credit card (c)} is required, exactly one security policy between {\tt high (h)} and {\tt standard (s)} must be chosen, with the constraint that {\tt c} implies {\tt h}.
Suppose that $\nu$ gives value $2$ to literal $b$ and to literal $c$, value $1$ to literal $h$, and
value $0$ to every other literal and that $(\mathbb{N}, +, <)$ is the monotone, totally ordered semigroup under consideration.

$C$ has $4$ models. The models of $C$ are reported in the following table and for each of them, we indicate its value
according to $\nu$.
\begin{center}
\begin{tabular}{cc}
\toprule
$(b, c, h, s) \in \valida(C)$ & $\nu(b, c, h, s)$\\
\midrule
$(0, 1, 1, 0)$ & $3$\\
$(1, 0, 0, 1)$ & $2$\\
$(1, 0, 1, 0)$ & $3$\\
$(1, 1, 1, 0)$ & $5$\\
\bottomrule
\end{tabular}
\end{center}
For this example, the set of top-$2$ values is $\{5, 3\}$. 
There are two possible sets of top-$2$ solutions, namely $\{(1, 1, 1, 0), (0, 1, 1, 0)\}$ 
and $\{(1, 1, 1, 0), (1, 0, 1, 0)\}$. Any circuit in the \dDnnf\ language over $X = \{b, c, h, s\}$ having as models
$\{(1, 1, 1, 0), (0, 1, 1, 0), (1, 0, 1, 0)\}$ is an admissible result for the top-$2$ transformation of $C$.
\end{example}

%
%
%

\section{Computing Top-$k$ Solutions}\label{sec:topsol}

For keeping the 
presentation simple enough, we assume in the following 
that the \dDnnf\ circuits that are considered as inputs satisfy a few assumptions. First, we suppose that every internal node in such a circuit (whatever it is
a $\wedge$ node or a $\vee$ node) is binary (it has two children). We also assume that the circuit is smooth,
which means that the sets of variables associated with the two children of any $\vee$ node are the same one. We finally assume that
those \dDnnf\ circuits are reduced, in the sense that no leaf node labelled by a Boolean constant occurs in the circuit, unless
the circuit is such a leaf (in which case the optimization tasks trivialize). 
Those three assumptions are computationally harmless: any \dDnnf\ circuit can be binarized in linear time (every internal $N$ node having $m > 2$
children $N_1, \ldots, N_m$ can be replaced by a binary tree with $m-1$ internal nodes of the same type as $N$ and $N_1, \ldots, N_m$  as children; and every internal node with a single child can be replaced by its child), smoothed in quadratic time \cite{darwiche01tractable} (and even more efficiently for structured \dDnnf\ \cite{DBLP:conf/nips/ShihBBA19}), and reduced in linear time (just applying the elementary rules of Boolean calculus). 
Another reasonable assumption concerns the representation of the values in $K$ and the costs of computing $\otimes$ and of comparing
elements of $K$ using $<$: one supposes that the size of the representations of the values in $K$ are bounded by a constant and
that $\otimes$ and $<$ are constant-time operations. 

We have obtained the following result:
 
\begin{theorem}
\label{th:topkconfddnnf}
Let $(K,\otimes, <)$ be a monotone, totally ordered semigroup. Let $X$ be a set of variables. Let $\nu$ be a value function over $X$ onto $K$. Let $C$ be a \dDnnf\ circuit over $X$. The problem $\topconf{k}(\nu,C)$ can be solved in time $O(|C| \cdot k \cdot log~k)$. 
\end{theorem}

\begin{algorithm}
\KwIn{$N$: a node in a \dDnnf\  circuit, $k$: a positive integer, $\nu$: the value function}
\KwResult{a list of top-$k$ solutions of the \dDnnf\  circuit rooted at $N$}
\If{$N$ is a leaf node labelled by literal $\ell$}{
     \Return [[$\ell$]]\;
}
top\_c0 = $\topconf{k}(\nu, N.children(0))$\;
top\_c1 = $\topconf{k}(\nu, N.children(1))$\;
\eIf(){$N$ is a $\vee$ node}{
    \Return sorted\_fusion(top\_c0, top\_c1, k, $\nu$)\;
}(~~\tcp*[h]{$N$ is a $\wedge$ node}){
    \Return sorted\_product(top\_c0, top\_c1, k, $\nu$)\;
}
\caption{$\topconf{k}(\nu,N)$.\label{alg:topk}}
\end{algorithm}

\paragraph{An algorithm for the top-$k$ solutions problem.}
As a constructive proof of this theorem, we present Algorithm \ref{alg:topk} that solves $\topconf{k}(\nu,C)$ within the expected amount of time. 
This algorithm computes in a bottom-up fashion the values of two synthesized attributes (representing top-$k$ solutions and 
their values) for the \dDnnf\ circuits rooted at the nodes of the input circuit $C$.  When $N$ is an internal node of $C$, 
top\_c0 (resp. top\_c1) denotes the list of top-$k$ solutions (and the corresponding values for $\nu$)
that has been computed for the \dDnnf\ circuit rooted at the left (resp. right) child of $N$.
Depending on the label of $N$ ($\wedge$ or $\vee$), a different procedure is run to derive
the values of the two attributes at $N$: sorted\_fusion(top\_c0, top\_c1, k, $\nu$) when $N$ is a $\vee$ node 
and sorted\_product(top\_c0, top\_c1, k, $\nu$) when $N$ is a $\wedge$ node. 
When called at a $\vee$ node, sorted\_fusion returns the ordered list of top-$k$ solutions extracted from the sorted union 
of top\_c0 and top\_c1. When called at a  $\wedge$ node, sorted\_product returns an ordered list of top-$k$ solutions 
generated from the cross product of top\_c0 and top\_c1.

\paragraph{Correctness and complexity.}
In order to prove the correctness of our bottom-up algorithms for computing $k$ top solutions 
of a \dDnnf\ circuit $C$, one must show that at any internal node $N$ of $C$ computing $k$ top solutions 
of the \dDnnf\ circuit $C_N$ rooted at $N$ can be achieved when $k$ top solutions 
of the \dDnnf\ circuits $C_{N_0}$ and $C_{N_1}$, rooted respectively at the children $N_0$ and $N_1$ of $N$, 
have been computed first. Note that each of $C_{N_0}$ and $C_{N_1}$ has at least one model since $C$ is supposed
to be simplified.

The expected result is rather obvious when $N$ is a $\vee$ node: in this case, by definition, the set of valid assignments of the circuit 
rooted at $N$ is the union of the sets of valid assignments of the circuits rooted at $N_0$ and at $N_1$. 
Thus, to derive $k$ top solutions of the \dDnnf\ circuit rooted at $N$, it is enough to make the union
of the sets of $k$ top solutions associated with its two children, to sort them in decreasing order 
w.r.t. their values, and to keep the first $k$ elements of the sorted list of solutions. 

Things are a bit more tricky when considering $\wedge$ nodes $N$. In that case, the monotony assumption
about $\otimes$ is useful. 
Towards a contradiction, suppose that the set $\{\omega \cdot \omega' : \omega \in \topconf{k}(\nu,C_{N_0}), \omega \in \topconf{k}(\nu,C_{N_1})\}$
does not include $k$ top solutions of $C_N$. Thus, there exist $\omega \models C_{N_0}$ and $\omega' \models C_{N_1}$
such that $\omega \not \in \topconf{k}(\nu,C_{N_0}) = \{\omega_1, \ldots, \omega_k\}$ and for each $i \in [k]$,
$\nu(\omega \cdot \omega') > \nu(\omega_i \cdot \omega'_1)$, where $\omega'_1 \in \topconf{1}(\nu,C_{N_1})$.
In such a case, since $\nu(\omega') \leq \nu(\omega'_1)$ and $\nu(\omega) \leq \nu(\omega_i)$ ($i \in [k]$),
the monotony of the semigroup ensures that for each $i \in [k]$, $\nu(\omega) \otimes \nu(\omega') \leq \nu(\omega_i) \otimes \nu(\omega'_1)$,
or equivalently $\nu(\omega \cdot \omega') \leq \nu(\omega_i \cdot \omega'_1)$, a contradiction.

\bigskip

As to complexity, let us first consider a simple, yet naive implementation of the two functions sorted\_fusion and sorted\_product.
Function sorted\_fusion(top\_c0, top\_c1, k, $\nu$) computes the sorted union of two disjoint lists of size $k$. 
It is well-known that this can be done in time $O(k \cdot \log~k)$.
Function  sorted\_product(top\_c0, top\_c1, k, $\nu$) can be implemented by computing explicitly the cross 
product of the two lists of $k$ solutions, ordering them and picking up top-$k$ elements.  
Through this implementation, the algorithm runs in time $O(|C| \cdot k^2 \cdot \log~k \cdot |X| )$.

However, a better implementation can be obtained by avoiding to sort at each $\wedge$ node $N$ 
all the assignments resulting from the cross product of the two lists, top\_c0 and top\_c1. 
Our algorithm takes advantage of a max-heap implementation of a priority queue $Q$, 
i.e., a data structure allowing to add elements to $Q$ and remove elements from $Q$ in time logarithmic in the size of $Q$, 
and also to retrieve an element of maximal value from $Q$ in constant time.
In our implementation of sorted\_product, one stores in $Q$ pairs of indices $(i,j)$ together with the corresponding value for $\nu$ 
denoted by $\nu(i, j)$.  Each pair $(i, j)$ represents the concatenation of the $i$th assignment 
of top\_c0 with the $j$th assignment of top\_c1.
Because  $(K,\otimes, <)$ is monotone, the first pair $(0, 0)$ of $Q$ is one of top value. Then the following treatment is iterated for $k-1$ steps.
At each step, the first pair stored in $Q$, $(i,j)$ is retrieved, then deleted from $Q$, and the corresponding assignment is added to the list
of top-$k$ solutions under construction for node $N$.
Then the pairs  $(i,j+1)$ and $(i+1,0)$ are added to the queue if they were not added yet. 
We use additional structures to check these in a decent time (this test can be done in constant time using a hashmap or in time $\log~k$ using B+ trees).
By construction, after $k$ steps, the list at node $N$ contains $k$ top assignments of the cross product between top\_c0 and top\_c1.
Since at each step, one element is removed from $Q$ and at most two elements are added, the size of $Q$ increases linearly in the number of steps and therefore the size of $Q$ remains linear in $k$. 
Therefore, the time complexity of computing $k$ top assignments for a $\wedge$ node $N$ is in 
$O(k \cdot \log k   \cdot |X|)$.

The $|X|$ factor in the complexity evaluation comes from the computational cost of concatenating the two assignments represented in a naive manner.
It is possible to remove this factor through a more efficient representation of assignments. 
In our implementation, sets of literals are represented as binary trees where leaves are labelled by literals. Then, the concatenation 
of partial assignments at $\wedge$ nodes can be achieved in constant time  by taking the roots of the two sets and creating a new root having them as children. The decomposability of $\wedge$ nodes ensures the correctness of the approach (i.e., the resulting tree is guaranteed to correspond to a partial assignment).
At $\vee$ nodes, where unions of sets of partial assignments must be done, there is no need for equality tests to avoid duplicates: by construction, 
the determinism of $\vee$ nodes ensures that those unions are disjoint ones. As a consequence, the construction of solutions can be done efficiently through the tree 
representation of assignments and the multiplicative factor $|X|$ can be removed from the time complexity of our algorithm.

Finally, the time used by sorted\_fusion(top\_c0, top\_c1, k, $\nu$) can be improved given that top\_c0 and top\_c1 are sorted. It is well-known that sorting two sorted lists can be done in time linear in the sum of the sizes of the lists. This has no impact on Theorem \ref{th:topkconfddnnf} but in practice, it leads to significant time savings.

\begin{example}
Let us consider Example \ref{ex:ex1} again. The nodes $N$ of the DAG reported in Figure \ref{fig:ddnnf:topkconf} correspond in a bijective way with
those of the \dDnnf\ circuit in Figure \ref{fig:ddnnf:ddnnf} (the nodes and the arcs of the two DAGs are the same ones, only the labels change). 
The label of each node $N$ of  the DAG at Figure \ref{fig:ddnnf:topkconf} is a list of top-$2$ solutions of the \dDnnf\ circuit in Figure \ref{fig:ddnnf:ddnnf}
rooted at the same node (for the sake of readability, the values of those solutions are not reported on the figure).
\end{example}

\section{Computing Top-$k$ Values}\label{sec:topval}

From the user perspective, computing top-$k$ solutions requires first to decide which value of $k$ should be retained.
To make an informed choice, deriving first top-$k$ values (with possibly another value for $k$ than the one representing the number
of solutions) can be very useful. 
Indeed, to make things simple, suppose that the top-$5$ values for a given scenario are $100, 99, 98, 10, 2$. Here, there is a huge gap between the first top-$3$ values and the two remaining ones. The user can then be tempted to ask
first for the computation of top-$3$ solutions, and then look at their values. If the values are respectively $100, 99, 98$, he/she
may decide to stop the computation because he/she knows that there are only one top-$1$ solution, and only two top-$2$
solutions and he/she is fine with them. Contrastingly, if the computed values are respectively $100, 100, 100$, he/she may ask for
the computation of more top solutions. 

To deal with this issue, we have designed a pseudo polynomial-time algorithm for solving the top-$k$ values problem, provided that the semigroup $(K,\otimes, <)$ is almost strictly monotone. Though more demanding than the monotony condition, this restriction is met by several semigroups that are useful for modeling utilities or probabilities (as sketched in Section \ref{sec:topk}). We have obtained the following result: 

\begin{theorem}
\label{th:strictvalue}
Let $(K,\otimes, <)$ be an almost strictly monotone, totally ordered semigroup. Let $X$ be a set of variables. Let $\nu$ be a value function over $X$ onto $K$. Let $C$ be a \dDnnf\ circuit over $X$.  The problem $\topval{k}(\nu,C)$ can be solved in time $O(|C| \cdot k^2 \cdot log~k)$. 
\end{theorem}

\paragraph{An algorithm for the top-$k$ values problem.} 
Our algorithm to solve $\topval{k}(\nu,C)$ is a variant of Algorithm \ref{alg:topk} for computing top-$k$ solutions.
A main difference is that it is sufficient to store values and thus, the procedures 
sorted\_fusion and  sorted\_product take as inputs tables of values and not tables of pairs (assignment, value) and they output tables of values.
Those procedures must be updated to handle (respectively) $\vee$ nodes and $\wedge$ nodes in a satisfying way since it is
possible in the top-$k$ values context that duplicates appear. When computing top-$k$ solutions,
the decomposability and the determinism conditions on \dDnnf\ circuits ensure that the assignments generated 
at each node of $C$ when applying Algorithm \ref{alg:topk} are distinct, but it is not the case for values.

Thus, when calling sorted\_fusion(top\_c0, top\_c1, k, $\nu$), it may happen that the same value appears both in top\_c0 and top\_c1 so that one of the duplicates has to be removed after sorting. Thus the update of sorted\_fusion(top\_c0, top\_c1, k, $\nu$) simply consists in sorting the values of the union of top\_c0 and top\_c1 and then removing the duplicates from the resulting sorted table. This has no impact on the complexity of sorted\_fusion. 
When calling sorted\_product(top\_c0, top\_c1, k, $\nu$) at a $\wedge$ node,
it is also possible to get duplicates as distinct $\otimes$-combinations of the values of the solutions of its two children. 
The update of sorted\_product(top\_c0, top\_c1, k, $\nu$) can be done as follows. The values that are obtained are kept in memory, using a binary search tree $S$ allowing us to add an element and to check whether an element is already stored in $S$ in time logarithmic in the size of $S$. Whenever we pop a pair $(i,j)$, we check using $S$ whether $\nu(i,j)$ has already been outputted. $\nu(i,j)$ is then outputted and added to $S$
iff it has not been outputted before.
The remaining instructions of the algorithm, i.e., adding $(i+1,0)$ and $(i,j+1)$ to $Q$, are the same ones as those in the algorithm for computing top-$k$ solutions.
This treatment is repeated until $k$ distinct values have been found or we went through all the pairs $(i,j)$.

\paragraph{Correctness and complexity.}
The main point for proving the correctness of our algorithm is the correctness of sorted\_product. For this, we can prove that if a value $v$ belongs to the top-$k$ values of a subcircuit rooted at a $\wedge$ node $N$ with children $N_0$ and $N_1$, then either $v$ is equal to the least absorptive element of$K$ if it exists or there exist $u$ in the top-$k$ values of $N_0$ and $w$ in the top-$k$ values of $N_1$ such that $u \otimes w = v$. This property is a consequence of the fact that 
$(K,\otimes,<)$ is almost strictly monotone.

\bigskip
The time complexity of our algorithm comes from doing $|C|$ times a call to the procedure 
sorted\_fusion(top\_c0, top\_c1, k, $\nu$) or to the procedure sorted\_product(top\_c0, top\_c1, k, $\nu$). For sorted\_fusion, the complexity bound is the same one as for the top-$k$ solutions case. For  sorted\_product, one may need to consider the full set of $k^2$ pairs of values coming from the $\otimes$-combinations of the top-$k$ values associated with the children of the $\wedge$ node at hand. Therefore, the time complexity of a call to this procedure is in $O(k^2 \cdot \log k)$.

\begin{example}
Let us step back to Example \ref{ex:ex1} once more. The nodes $N$ of the DAG reported in Figure \ref{fig:ddnnf:topkval} correspond in a bijective way with
those of the \dDnnf\ circuit in Figure \ref{fig:ddnnf:ddnnf}. The label of each node $N$ of the DAG reported in Figure \ref{fig:ddnnf:topkval} is the list of top-$2$ values of the \dDnnf\ circuit in Figure \ref{fig:ddnnf:ddnnf} rooted at the corresponding node.
\end{example}

\section{Top-$k$ Transformation}\label{sec:toptra}

Interestingly, the top-$k$ transformation problem can also be solved in polynomial time
when $C$ is a \dDnnf\ circuit provided that the totally ordered semigroup at hand is almost strictly monotone.

\begin{theorem}
\label{th:transformation}
Let $(K,\otimes, <)$ be an almost strictly monotone, totally ordered semigroup. 
Let $X$ be a set of variables. Let $\nu$ be a value function over $X$ onto $K$. Let $C$ be a \dDnnf\ circuit over $X$. 
The problem  $\transfv{k}(\nu,C)$ can be solved in time $O(|C| \cdot k^2 \cdot log~k)$ and the resulting \dDnnf\ circuit has a size in $O(|C| \cdot k^2)$.

\end{theorem}

\paragraph{An algorithm for the top-$k$ transformation problem.}
Our algorithm for $\transfv{k}(\nu,C)$ starts by running $\topval{k}(\nu,C)$ so that the top-$k$ values of the \dDnnf\ circuits
rooted at the nodes $N$ of $C$ are stored as additional labels of the corresponding nodes.  
The generation of a \dDnnf\ circuit $C'$ as a result of $\transfv{k}(\nu,C)$ is achieved by parsing the 
nodes $N$ of $C$ (together with the list $L_N$ of values associated with them) in a bottom-up manner.
For the sake of simplicity, let us suppose first that $(K,\otimes, <)$ does not have a least absorptive element.
Let $N_0$ and $N_1$ be the children of $N$ when $N$ is an internal node.
The treatment is as follows. Every leaf node of $C$ is kept unchanged.
For each internal node $N$ together with the associated list $L_N$ of values, 
we create one new node for each value $v$ in the list $L_N$. The node created from $N$ with value $v$ is noted $(N, v)$.

If $N$ is a $\vee$ node, then  $(N, v)$ is a $\vee$ node, and 
for each value $v$,  arcs connecting $(N, v)$ to $(N_0,v)$ and/or $(N_1,v)$ are added if those last nodes exist, 
i.e., if $v$ belongs to the list of the top-$k$ values associated with $N_0$ and/or $N_1$.
If the root $N$ of $C$ is a $\vee$ node, then the root of $C'$ is a new $\vee$ node having as children the nodes $(N, v)$ 
where $v$ varies in $L_N$.
Now, for a $\wedge$ node $N$, for each value $v$ in the list of the top-$k$ values associated with $N$, 
let $L(v)$ be the list of pairs $(u,w)$ of values from the lists of the top-$k$ values associated respectively with $N_0$ and 
$N_1$, such that $u \otimes w = v$. 
We construct a subcircuit rooted at $(N,v)$ that encodes the disjunction over the values $v$ of the conjunctions of 
$(N_0,u)$ and $(N_1,w)$ such that $u \otimes w = v$.

Once the root of $C$ has been processed, the resulting circuit is simplified in linear time by removing every node and every arc that cannot
be reached from the root, and by shunting every node that has a single child.

When the semigroup has a least absorptive element $a$, the construction of $C'$ is similar to the previous one, except when $a$ belongs to the top-$k$ values of a subcircuit rooted at a $\wedge$ node $N$. Indeed, in such a case, $a$ also belongs to the top-$k$ values associated with one of the children $N_0$, $N_1$ of $N$. Suppose that $a$ is a top-$k$ value associated with $N_0$. Because $a$ is absorptive, for any valid assignment $\omega$ of the subcircuit rooted at $N_1$ such that $\nu(\omega) = v$ (whatever $v$ is), we have $a \otimes v = a$. Therefore, $\omega$ can be among the valid assignments of the subcircuit rooted at $N_1$ that produce (once concatenated with a valid assignment of the subcircuit rooted at $N_0$) a top-$k$ solution at node $N$. Accordingly, all the valid 
assignments $\omega$ at $N_1$ must be stored, even if they are not among those having a top-$k$ value at $N_1$.  To deal with this case, 
we copy $C$ into a new circuit $C^c$ so that for each node $N$ of $C$ we have an associated node in $C^c$, denoted $N^c$.
The subcircuit associated with each node $(N,v)$ where $v$ is different of $a$  
is built as explained before.
We now explain how to build the subcircuits associated with the nodes $(N,a)$.

The construction depends on the type of $N$:
\begin{itemize}
\item  if $N$ is a $\vee$ node, then the subcircuit associated with $(N,a)$ is equal to the disjunction of the nodes $(N_0,a)$ and $(N_1,a)$ if they exist. If there exists only one such node, $(N,a)$ is equal to this node. 
\item if $N$ is a $\wedge$ node, then there are three cases: if both $(N_0,a)$ and $(N_1,a)$ exist, then the subcircuit associated with $(N,a)$ is the disjunction of the conjunction of $(N_0,a)$ and $N_1^c$ with the conjunction of $N_0^c$ and $(N_1,a)$; if only $(N_0,a)$ exists, then the subcircuit associated with $(N,a)$ is the 
conjunction of $(N_0,a)$ and $N_1^c$; in the remaining case, i.e., when $(N_1,a)$ exists, then the subcircuit associated with $(N,a)$ is the conjunction of $N_0^c$ and $(N_1,a)$.

\end{itemize}

\paragraph{Correctness and complexity.}
By construction, the decomposability of the $\wedge$ nodes is preserved by the transformation algorithm. Furthermore, the $\vee$ nodes that are created in the resulting circuit are deterministic ones (each child of such a node corresponds to a set of satisfying assignments
having a value different of those of its sibling or obtained by concatenating distinct partial assignments, therefore those sets of assignments are pairwise disjoint).

The main part of the correctness of the algorithm is a generalization of the property used for proving the correctness of our algorithm for solving $\topval{k}(\nu,C)$.
Let $N$ be a $\wedge$ node with children $N_0$ and $N_1$. Let $v$ be a value among the top-$k$ values of the circuit rooted at $N$. Then, if there exist $u$ and $w$ values of assignments from $N_1$ and $N_2$ such that $u \times w =  v$ then $v$ is the absorptive element or $u$ and $v$ are in the top-$k$ values of $N_1$ and $N_2$.

\bigskip

The multiplicative factor $k^2$ in the size of the resulting \dDnnf\ circuit comes directly from the fact that in the worst case, 
every $L_N$ contains $k^2$ elements.

\begin{example}
Let us consider again Example \ref{ex:ex1} once more. Figure \ref{fig:toptra:algo} illustrates the computation achieved for deriving $\transfv{2}(\nu,C)$
where $C$ is the \dDnnf\ circuit presented at Figure \ref{fig:ddnnf:ddnnf}. An equivalent, yet simplified circuit is reported on Figure \ref{fig:toptra:result}.
\end{example}

\newcommand{\topTraFigNodeFadeColor}{black!60}

\newcommand{\topTraFigNode}[4]{
  \node[topTraFigNodeStyle] (#1) at (#2) {[#4] #3};
}
\newcommand{\topTraFigOneNode}[4]{
  \node[topTraFigNodeStyle] (#1) at (#2) {[#4] #3};
  \node[topTraFigBoxStyle] () at (#2) {};
}
\newcommand{\topTraFigOneNodeFade}[4]{
  \node[topTraFigNodeFadeStyle] (#1) at (#2) {[#4] #3};
  \node[topTraFigBoxFadeStyle] () at (#2) {};
}
\newcommand{\topTraFigTwoNodes}[9]{
  \node[topTraFigNodeStyle] (#1) at (#2) {[#4] #3};
  \node[topTraFigNodeStyle] (#5) at (#6) {[#8] #7};
  \node[topTraFigDblBoxStyle] () at (#9) {};
}
\newcommand{\topTraFigTwoNodesFadeLower}[9]{
  \node[topTraFigNodeStyle] (#1) at (#2) {[#4] #3};
  \node[topTraFigNodeFadeStyle] (#5) at (#6) {[#8] #7};
  \node[topTraFigDblBoxStyle] () at (#9) {};
}

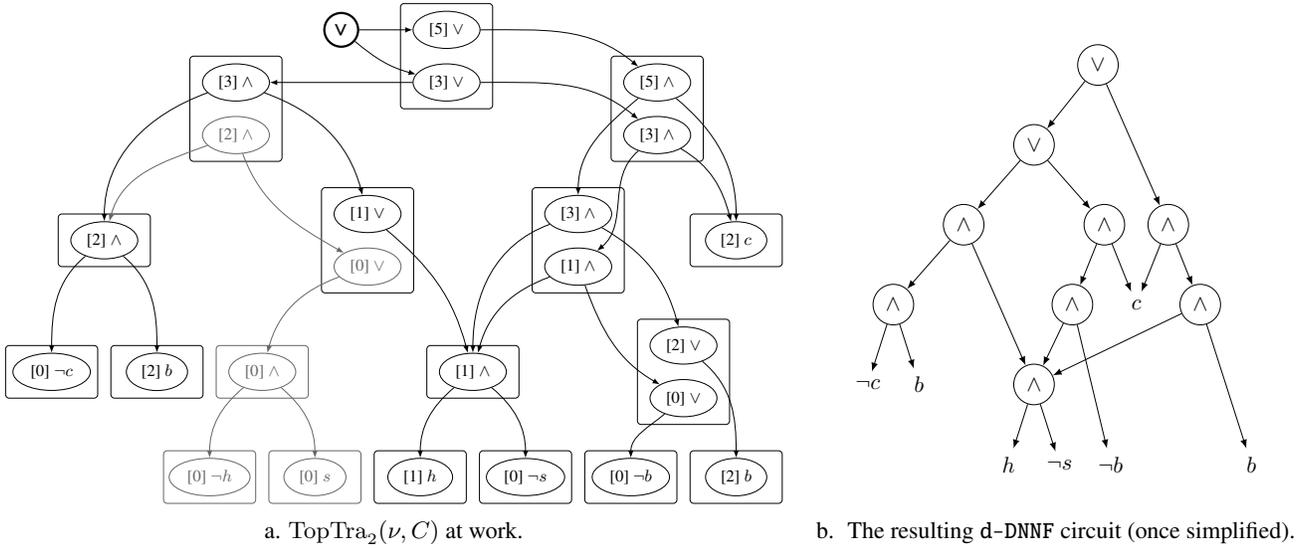
\begin{figure*}[t]
  \centering
  \begin{subfigure}[b]{0.59\textwidth}
    \centering
    \scalebox{.7}{
    \begin{tikzpicture}
      \usetikzlibrary{shapes.geometric}
        \tikzstyle{topTraFigNodeStyle}=[align=left, font=\small, draw, ellipse]
        \tikzstyle{topTraFigNodeFadeStyle}=[topTraFigNodeStyle, draw=\topTraFigNodeFadeColor, text=\topTraFigNodeFadeColor]
        \tikzstyle{topTraFigBoxStyle}=[draw, rectangle, rounded corners=2pt, minimum width=1.75cm, minimum height=1cm]
        \tikzstyle{topTraFigBoxFadeStyle}=[topTraFigBoxStyle, draw=\topTraFigNodeFadeColor]
        \tikzstyle{topTraFigDblBoxStyle}=[topTraFigBoxStyle, minimum height=2cm]
        \tikzstyle{link}=[->,>=latex]
        \tikzstyle{linkFade}=[->,>=latex,color=\topTraFigNodeFadeColor]
        
        \topTraFigOneNodeFade{nh}{0,0.5}{$\neg h$}{0}
        \topTraFigOneNodeFade{s}{2,0.5}{$s$}{0}
        \topTraFigOneNode{h}{4,0.5}{$h$}{1}
        \topTraFigOneNode{ns}{6,0.5}{$\neg s$}{0}
        \topTraFigOneNode{nb}{8,0.5}{$\neg b$}{0}
        \topTraFigOneNode{b}{10,0.5}{$b$}{2}

        \topTraFigOneNode{nc}{-3,2.5}{$\neg c$}{0}
        \topTraFigOneNode{b2}{-1,2.5}{$b$}{2}
        \topTraFigOneNodeFade{and1}{1,2.5}{$\wedge$}{0}
        \draw[linkFade] (and1) edge[out=220,in=90] (nh);
        \draw[linkFade] (and1) edge[out=320,in=90] (s);
        \topTraFigOneNode{and2}{5,2.5}{$\wedge$}{1}
        \draw[link] (and2) edge[out=220,in=90] (h);
        \draw[link] (and2) edge[out=320,in=90] (ns);        
        \topTraFigTwoNodes{or11}{9,3}{$\vee$}{2}{or12}{9,2}{$\vee$}{0}{9,2.5}
        \draw[link] (or11) edge[out=320,in=90] (b);
        \draw[link] (or12) edge[out=220,in=90] (nb);

        \topTraFigOneNode{and3}{-2,5}{$\wedge$}{2}
        \draw[link] (and3) edge[out=220,in=90] (nc);
        \draw[link] (and3) edge[out=320,in=90] (b2);
        \topTraFigTwoNodesFadeLower{or21}{3,5.5}{$\vee$}{1}{or22}{3,4.5}{$\vee$}{0}{3,5}
        \draw[link] (or21) edge[out=320,in=105] (and2);
        \draw[linkFade] (or22) edge[out=200,in=75] (and1);
        \topTraFigTwoNodes{and41}{7,5.5}{$\wedge$}{3}{and42}{7,4.5}{$\wedge$}{1}{7,5}
        \draw[link] (and41) edge[out=200,in=90] (and2);
        \draw[link] (and41) edge[out=330,in=105] (or11);
        \draw[link] (and42) edge[out=200,in=75] (and2);
        \draw[link] (and42) edge[out=290,in=150] (or12);
        \topTraFigOneNode{c}{10,5}{$c$}{2}

        \topTraFigTwoNodesFadeLower{and51}{.5,8}{$\wedge$}{3}{and52}{.5,7}{$\wedge$}{2}{.5,7.5}
        \draw[link] (and51) edge[out=200,in=90] (and3);
        \draw[link] (and51) edge[out=340,in=105] (or21);
        \draw[linkFade] (and52) edge[out=200,in=75] (and3);
        \draw[linkFade] (and52) edge[out=290,in=150] (or22);
        \topTraFigTwoNodes{and61}{8.5,8}{$\wedge$}{5}{and62}{8.5,7}{$\wedge$}{3}{8.5,7.5}
        \draw[link] (and61) edge[out=220,in=90] (and41);
        \draw[link] (and61) edge[out=320,in=90] (c);
        \draw[link] (and62) edge[out=220,in=40] (and42);
        \draw[link] (and62) edge[out=330,in=105] (c);

        \topTraFigTwoNodes{or31}{4.5,9}{$\vee$}{5}{or32}{4.5,8}{$\vee$}{3}{4.5,8.5}
        \draw[link] (or31) edge[out=0,in=140] (and61);
        \draw[link] (or32) edge[out=0,in=140] (and62);
        \draw[link] (or32) edge[out=180,in=0] (and51);
        \node[draw, circle, very thick] (root) at (2.5,9) {$\pmb{\vee}$};
        \draw[link] (root) edge[] (or31);
        \draw[link] (root) edge[out=320,in=165] (or32);
    \end{tikzpicture}
    }
    \caption{\label{fig:toptra:algo}$\transfv{2}(\nu,C)$ at work.}
  \end{subfigure}
  \begin{subfigure}[b]{0.39\textwidth}
    \centering
    \scalebox{\figScale}{
    \begin{tikzpicture}
      \tikzstyle{internalAndNode}=[draw, circle]
      \tikzstyle{internalOrNode}=[draw, circle]
      \tikzstyle{leafNode}=[]
      \tikzstyle{link}=[->,>=latex]

      \node[leafNode] (h) at (0, 0) {$h$};
      \node[leafNode] (ns) at (.8, 0) {$\neg s$};
      \node[internalAndNode] (and1) at (.4,1.25) {$\wedge$};
      \draw[link] (and1) -- (h);
      \draw[link] (and1) -- (ns);

      \node[leafNode] (nc) at (-2.2, 1.25) {$\neg c$};
      \node[leafNode] (b) at (-1.4, 1.25) {$b$};
      \node[internalAndNode] (and2) at (-1.8,2.5) {$\wedge$};
      \draw[link] (and2) -- (nc);
      \draw[link] (and2) -- (b);

      \node[internalAndNode] (and3) at (-.7,3.75) {$\wedge$};
      \draw[link] (and3) -- (and2);
      \draw[link] (and3) -- (and1);

      \node[leafNode] (nb) at (1.6, 0) {$\neg b$};
      \node[internalAndNode] (and4) at (1,2.5) {$\wedge$};
      \draw[link] (and4) -- (and1);
      \draw[link] (and4) -- (nb);

      \node[leafNode] (c) at (2, 2.5) {$c$};
      \node[internalAndNode] (and5) at (1.5,3.75) {$\wedge$};
      \draw[link] (and5) -- (and4);
      \draw[link] (and5) -- (c);

      \node[internalOrNode] (or1) at (.4,5) {$\vee$};
      \draw[link] (or1) -- (and3);
      \draw[link] (or1) -- (and5);

      \node[leafNode] (b2) at (3.8, 0) {$b$};
      \node[internalAndNode] (and6) at (3, 2.5) {$\wedge$};
      \draw[link] (and6) -- (and1);
      \draw[link] (and6) -- (b2);

      \node[internalAndNode] (and7) at (2.5, 3.75) {$\wedge$};
      \draw[link] (and7) -- (c);
      \draw[link] (and7) -- (and6);

      \node[internalOrNode] (root) at (1.4,6.25) {$\vee$};
      \draw[link] (root) -- (or1);
      \draw[link] (root) -- (and7);

      \node[]() at (0,-.5) {}; 
      
    \end{tikzpicture}
    }
    \caption{\label{fig:toptra:result} The resulting \dDnnf\ circuit (once simplified).}
  \end{subfigure}
  \caption{\label{fig:toptra} $\transfv{2}(\nu,C)$ at work on Example \ref{ex:ex1}. Each box in Figure \ref{fig:toptra:algo} gathers nodes related to the same node in the initial \dDnnf\ circuit $C$ given at Figure \ref{fig:ddnnf:ddnnf}. Each node is labelled by the value given to it according to $\topval{2}(\nu,C)$ and by a connective or a literal. Faded nodes and arcs are reported to illustrate the construction but they are not parts of the  \dDnnf\ circuit that is generated. 
 Figure \ref{fig:toptra:result} presents the resulting \dDnnf\ circuit (the circuit given at Figure \ref{fig:toptra:algo}, once simplified).}
\end{figure*}

Interestingly, our transformation algorithm leads to the generation of a \dDnnf\ circuit where all the assignments satisfying a subcircuit rooted at a $\wedge$ node
have the same values. This property opens the door for a number of additional tractable treatments that could be useful, like
counting the number of valid assignments having a given value among the top-$k$ ones, or uniformly sampling such assignments
using results of \cite{SGRM18}.


\section{Experimental Results}\label{sec:expe}

Clearly, all the top-$k$ algorithms presented in the previous sections prove practical when $k$ is small enough
(they run in linear time in the size of the \dDnnf\ circuit $C$ when $k$ is bounded by a constant). Notably, considering that $k$ is small enough is a reasonable assumption
since the generation of top-$k$ solutions is typically triggered by a human user who will not be able to encompass
a large set of solutions as a whole due to his/her cognitive limitations (see e.g., \cite{Miller56}).
However, the efficiency of our approach deeply relies on
the assumption that the constraints considered at start have been compiled into a \dDnnf\ circuit, and it is well-known that such a compilation step
can be computationally expensive (the size of the resulting \dDnnf\ circuit can be exponential in the size of the input constraints).

\paragraph{Empirical protocol.}

In order to figure out the benefits that can be reached by taking advantage of such a compilation-based approach, we
focused on the strictly monotone, totally ordered semigroup $(\mathbb{N}, +, <)$ and, in this setting, implemented 
an approach to the computation of top-$k$ solutions based on an algorithm for the {\sf NP}-hard problem called \textsc{Weighted Partial MaxSAT} problem.
As evoked previously, the semigroup $(\mathbb{N}, +, <)$ is suited to the (quite general) class of scenarios where the values under consideration
represent utilities.

Let us recall that an instance of \textsc{Weighted Partial MaxSAT} consists 
of a pair  $(C_{\mathrm{soft}}, C_{\mathrm{hard}})$ where $C_{\mathrm{soft}}$ and  $C_{\mathrm{hard}}$ are (finite) sets of weighted clauses,
and a weighted clause is an ordered pair $(c, w)$ where $w$ is a natural number or $\infty$.
Intuitively, $w$ gives the cost of falsifying $c$. If $w$ is infinite, the clause is hard, otherwise it is soft.
The objective is to determine a truth assignment that maximizes the sum of the weights of the 
clauses $c$ in $C_{\mathrm{soft}}$ that are satisfied, while satisfying all clauses $c$
such that $(c, \infty) \in C_{\mathrm{hard}}$. 
Now, starting with a \cnf\ formula $C$ over $X$ and a value function $\nu$ over $X$ onto 
$\mathbb{N}$, every clause $c$ of $C$ can be turned into a hard clause $(c, \infty) \in C_{\mathrm{hard}}$ and every literal $\ell$ over $X$
can be turned into a soft clause $(\ell, \nu(\ell)) \in C_{\mathrm{soft}}$. The truth assignment $\omega$ over $X$ that is obtained as a solution of 
$(C_{\mathrm{soft}}, C_{\mathrm{hard}})$ is by construction a top-$1$ solution of $C$ given $(\mathbb{N}, +, <)$ and $\nu$.
In order to leverage the approach so as to compute top-$k$ solutions, once $\omega$ has been generated, it is enough to add a hard clause equivalent to
$(\neg \omega, \infty)$ to $C_{\mathrm{hard}}$ in order to block the further generation of $\omega$ and to solve the resulting instance
of \textsc{Weighted Partial MaxSAT}. Once $k$ solutions have been generated (when this is possible), the procedure stops.

In order to compare the performances of our \dDnnf-driven algorithm for generating top-$k$ solutions with those of 
the \textsc{Weighted Partial MaxSAT}-based procedure sketched above for tackling the same issue, we made some experiments.
In our experimental evaluation, we took advantage of \dfour\footnote{\url{www.cril.univ-artois.fr/KC/d4.html}} \cite{LagniezMarquis17} to compile
\cnf\ formulae into  \dDnnf\ circuits. 
A Java library called \mm{Winston} has been developed.
It includes software for loading the \dDnnf\ circuit $C$ computed using \dfour, for smoothing it, and for computing top-$k$ solutions from it. 
For the top-$k$ solutions approach based on \textsc{Weighted Partial MaxSAT}, our procedure was empowered by one of the best solvers 
from the 2020 MaxSAT competition, namely {\tt MaxHS} \cite{10.1007/978-3-642-40627-0_21,10.1007/978-3-642-39071-5_13,davies2013solving}. 

We considered the dataset reported in \cite{LPAR-22:Knowledge_Compilation_meets_Uniform}. 
This dataset contains 1424 \cnf\ formulae coming from various fields, including probabilistic reasoning, bounded model checking, circuit, product configuration, SMTLib benchmarks, planning, quantified information flow and bug synthesis. 

We have run the compilation-based top-$k$ algorithm and the one based on {\tt MaxHS} for different values of $k$ ($1$, $5$, $10$, $20$) and, for each of these algorithms, we measured the time required to get $k$ solutions. 
For every literal $\ell$ over the variables $X$ of the input \cnf\ formula, an integer between $0$ and $\np{1000000}$ has been picked up
at random following a uniform distribution as the value of $\ell$. Following this approach, five value functions $\nu_1, \ldots, \nu_5$ have been generated per instance.
An instance has been viewed as solved when the corresponding algorithm for top-$k$ solutions succeeded in deriving $k$ top solutions for each of the five value functions before the timeout was reached.
For every instance solved, the mean time used to get $k$ top solutions when the value function varies has been considered.
For the compilation-based approach, the run time includes the time needed to compile the input \cnf\ formula into a \dDnnf\ circuit.  

The code of the algorithms and the data used in our experiments are available online.
All the experiments have been run on a cluster of computers based on bi-processors Intel Xeon E5-2680 v4 (2.2 GHz) with 768 GB of memory. For the experiments, we used a timeout set to 20 minutes.

\paragraph{Empirical results.}

\begin{figure}[t!]

\includegraphics[width=0.9\linewidth]{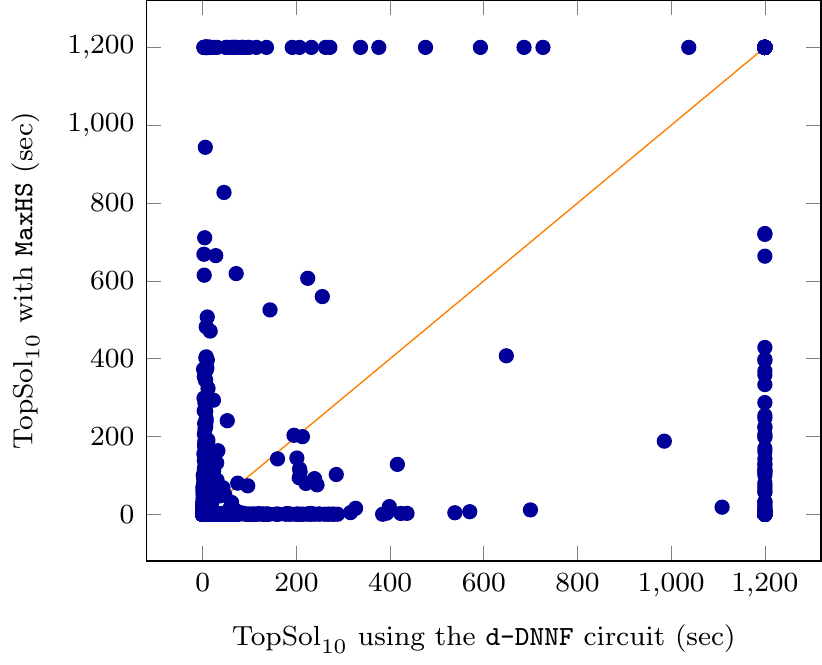}
\caption{Comparison of the runtimes of the compilation-based approach and the MaxSAT-based approach for computing $10$ top solutions.} 
\label{fig:result}
\end{figure}

The results are reported on the scatter plot at Figure~\ref{fig:result} and in Table \ref{table:topk}. 
Figure ~\ref{fig:result} shows a sharp separation about the run times required by the two approaches when $k=10$: the instances that 
were computationally easy for the approach based on {\tt MaxHS} were typically hard for compilation-based approach.
As reported in Table  \ref{table:topk}, the top-$k$ approach based on {\tt MaxHS} solved $94.0 \%$ of the instances of the dataset for $k=1$ to $87.2\%$ for $k=20$, 
while the compilation-based approach solved $81.5\%$ of the instances, whatever the value of $k$ among those considered in the experiments. 
When $k$ increased, the number of instances solved only by the MaxSAT-based approach diminished from $196$ for $k=1$ to $158$ for $k=20$,
while the number of instances solved only by the \dDnnf\ approach went up from $18$ to $78$.
As to run times, the fastest method was the MaxSAT-based approach for $k=1$, but around $k=10$ both approaches were tied, and for $k=20$, the compilation-based approach was clearly faster. This is not surprising since the main computational effort in the compilation-based approach is the time (and space) spent in the compilation phase, but this phase has to be performed once only, and the resources used are independent of the value of $k$. 
This contrasts with the MaxSAT-based approach to computing $k$ top solutions 
which requires to solve $k$ instances of an {\sf NP}-hard problem (the instances being possibly harder and harder).  

Accordingly, the compilation-based approach appears as the more interesting option to derive $k$ top solutions when $k$ is large enough.
Our experiments show that even for a small value of $k$, it can be a challenging method.

\begin{table}[ht]
\resizebox{\columnwidth}{!}{%
\begin{tabular}{lrrrr}
 \toprule
                                                      & k=1    & k=5    & k=10   & k=20   \\
 \midrule
   Success rate for {\tt MaxHS}                       & 94.0\% & 90.3\% & 88.6\% & 87.2\% \\
   Success rate for the \dDnnf\ approach              & 81.5\% & 81.5\% & 81.5\% & 81.5\% \\
   \# instances solved only by {\tt MaxHS}            & 196    & 170    & 165    & 158    \\
   \# instances solved only by the \dDnnf\ approach   & 18     & 44     & 65     & 78     \\
   \# instances where {\tt MaxHS} was faster          & 1154   & 897    & 637    & 544    \\
   \# instances where the \dDnnf\ approach was faster & 202    & 433    & 689    & 775    \\
 \bottomrule
 \end{tabular}
}
\caption{Performances of {\tt MaxHS} and of the \dDnnf\ approach for computing $k$ top solutions, depending on $k$.}
\label{table:topk}
\end{table}

Several limitations about our empirical protocol must be noted. For implementing the approach based on \textsc{Weighted Partial MaxSAT}, we took advantage of one solver only, namely {\tt MaxHS}. It would be interesting to determine whether the conclusions drawn would have been similar if another solver had been used instead.  
As to the compilation-based approach, the pipeline used to derive top-$k$ solutions is complex and it is based on different softwares implemented in different languages. The \dDnnf\ circuit computed using \dfour\ had to be saved on disk before being reloaded by the Java library, which took on average $1.1$s but went up to 47.5s.
The algorithm used in our experiments smoothes the \dDnnf\ circuit produced by \dfour, which is in practice quite a time-consuming operation (quadratic in the size of the circuit in the worst case). In our experiments it took on average $11$s and up to $410$s. 
We have not tried an implementation of the top-$k$ solutions algorithm that would be integrated in \dfour\ (so as to avoid the extra loading time) 
and would get rid of the smoothing operation. 

Despite these limitations, the compilation-based approach for the top-$k$ solutions problem appears as valuable. The bottleneck of this approach is typically the time 
required by \dfour\ to derive a \dDnnf\ circuit.
However, since the set of constraints representing the valid assignments is typically independent of the user, it has to be compiled once for all and the resulting circuit can be considered without any change for a number of users (each of them characterized by his/her own utility function $\nu$). This heavily contrasts with the approach to the top-$k$ solutions problem based on \textsc{Weighted Partial MaxSAT} where every choice of $\nu$ gives rise to a new instance of an {\sf NP}-hard problem that has to be solved. Finally, another advantage of the compilation-based approach is that it is not restricted to the semigroup $(\mathbb{N}, +, <)$ while \textsc{Weighted Partial MaxSAT} solvers rely on values that are numbers and that must be aggregated in an additive way.

\section{Discussion}\label{sec:disc}

Before concluding, we would like to mention that some of the restrictions considered in the previous sections could be questioned.
First, we have focused on the \dDnnf\ language mainly because existing compilers targeting the \Dnnf\ language,
including \adnan\ \cite{Darwiche01,Darwiche04}, \Dsharp\ \cite{Muiseetal12}, 
and \dfour\ (used in our experiments), actually target a subset of \dDnnf, namely the language of Decision-\Dnnf\ circuits.
Nonetheless, our top-$k$ algorithms could be extended to \Dnnf, i.e., removing the determinism condition on circuits $C$.
This would not have any impact on the complexity of the algorithms, especially those for computing the top-$k$ values and
making the top-$k$ transformation, and only a slight impact on the complexity of the algorithm for deriving $k$ top solutions
(equality tests should be implemented at $\vee$ nodes, thus one could not get rid of the $|X|$ factor in the complexity assessment).

The assumption according to which the size of the representations of the values in $K$ are bounded by a constant and that
$\otimes$ and $<$ are constant-time operations could also be relaxed. In that case, an extra factor in the time complexity of
the top-$k$ algorithms has to be added. This factor depends not only on the cost of performing the $\otimes$-operation over values
from $K$ represented using $m$ bits, but also of the size of the resulting value. For instance, summation combines two $m$-bits numbers
into an $m+1$-bit number in linear time, while product combines two $m$-bits numbers into a $2m$-bit number in
quadratic time -- if the naive schoolbook algorithm for multiplication is used. Accordingly, if $d$ is the depth of $C$,
the extra complexity factor to be considered is in $O(m+d)$ when $\otimes = +$, and this is not that much.
However, it is in $O(2^{2d} \cdot m^2)$ when $\otimes = \times$, which cannot be neglected when the
circuit $C$ is deep.

\section{Conclusion}\label{sec:concl}

We have presented three top-$k$ algorithms for \dDnnf\ circuits $C$ given a totally ordered, semigroup 
$(K, \otimes, <)$. Provided that some assumptions about monotony w.r.t. $<$ are satisfied by $\otimes$, 
these algorithms can be used, respectively, to compute in pseudo polynomial-time $k$ top solutions of $C$, 
the top-$k$ values met by solutions of $C$, and a \dDnnf\ circuit $C’$ satisfied exactly by the models of $C$ 
having a value among the top-$k$ ones. We have also presented the results of an empirical evaluation, showing 
that the \dDnnf\ compilation-based approach can prove valuable to address the top-$k$ solutions problem.

Among the perspectives for further research, we plan to extend our top-$k$ algorithms to a multicriteria setting, i.e.,
when several value functions $\nu_i$ onto distinct sets $K_i$ are considered at the same time. T
his would be useful to handle in a better way many applications about configuration, recommendation, and e-commerce. 
From the technical side, making such an extension would require a drastic 
update of the top-$k$ algorithms since the algebraic structure that underlies the present framework would not
be preserved.



\section*{Acknowledgements}
This work has been supported by the CPER DATA Commode project from the ``Hauts-de-France" Region.
It has also been partly supported by the PING/ACK project (ANR-18-CE40-0011) and the KOALA project (ANR-19-CE25-0003-01) from the French National Agency for Research.

\bibliographystyle{kr}
\bibliography{ddnnf}


\end{document}